\documentclass[journal]{IEEEtran}
\usepackage{amsmath,amsfonts}
\usepackage{algorithm}
\usepackage{multirow}
\usepackage{array}
\usepackage[caption=false,font=normalsize,labelfont=sf,textfont=sf]{subfig}
\usepackage{textcomp}
\usepackage{stfloats}
\usepackage{url}
\usepackage{svg}
\usepackage{hyperref}
\usepackage{verbatim}
\usepackage{graphicx}
\usepackage{cite}
\usepackage{xcolor}
\usepackage{microtype}
\usepackage{adjustbox}
\usepackage{algpseudocode}
\usepackage{array}
\usepackage{microtype}

\IEEEoverridecommandlockouts

\usepackage{booktabs}
\begin{document}

\title{HSIMamba: Hyperpsectral Imaging  Efficient  Feature Learning with Bidirectional State Space for Classification}

\author{{Judy X~Yang,~\IEEEmembership{Student Member,~IEEE},
        Jun~Zhou,~\IEEEmembership{Senior Member,~IEEE},
        Jing~Wang,~\IEEEmembership{Senior Member,~IEEE},
        Hui~Tian,~\IEEEmembership{Senior Member,~IEEE}, and~Alan Wee~Chung~Liew,~\IEEEmembership{Senior Member,~IEEE}}  
        
\thanks{Judy X Yang and Jun Zhou are with the School of Information and Communication Technology, Griffith University, Australia (corresponding author: Jun Zhou, jun.zhou@griffith.edu.au).}
\thanks{Jing Wang is with the Queensland Department of Agriculture and Fisheries, Australia.}}



\maketitle

\begin{abstract}
Classifying hyperspectral images (HSIs) is a difficult task in remote sensing, due to their complex high-dimensional data. To address this challenge, we propose HSIMamba, a novel framework that uses bidirectional reversed convolutional neural network (CNN) pathways to extract spectral features more efficiently. Additionally, it incorporates a specialized block for spatial analysis. Our approach combines the operational efficiency of CNNs with the dynamic feature extraction capability of attention mechanisms found in Transformers. However, it avoids the associated high computational demands.
HSIMamba is designed to process data bidirectionally, significantly enhancing the extraction of spectral features and integrating them with spatial information for comprehensive analysis. This approach improves classification accuracy beyond current benchmarks and addresses computational inefficiencies encountered with advanced models like Transformers. HSIMamba were tested against three widely recognized datasets - Houston 2013, Indian Pines, and Pavia University - and demonstrated exceptional performance, surpassing existing state-of-the-art models in HSI classification.
This method highlights the methodological innovation of HSIMamba and its practical implications, which are particularly valuable in contexts where computational resources are limited. HSIMamba redefines the standards of efficiency and accuracy in HSI classification, thereby enhancing the capabilities of remote sensing applications. Hyperspectral imaging has become a crucial tool for environmental surveillance, agriculture, and other critical areas that require detailed analysis of the Earth's surface. Please see our code in $\href{https://github.com/Judyxyang/judyxyang/blob/master/HSi_UH2013_P7_AB_VIM_V3_6_0330.ipynb}{HSI_Mamba}$ for more details.

\end{abstract}


\section{Introduction}

In remote sensing field,  hyperspectral imaging is the critical data source for land cover, containing hundreds of narrow wavelength bands spanning the entire electromagnetic spectrum, which can reflect the subtle earth surface~\cite{Cao2021,mateen2018role}. This technique allows for the precise identification and detection of materials, including those with spectral signatures that appear nearly identical in conventional visual representations, such as RGB images. The capacity to discern these subtle differences at a granular level holds significant promise for advanced Earth observation tasks. Applications range from detailed land cover mapping and precision agriculture to specific tasks such as detection~\cite{}, urban planning~\cite{tao2017automatic,vadrevu2022remote}, tree species classification~\cite{8529194}, and mineral exploration~\cite{shirmard2022review}. This capability underpins the value of hyperspectral imaging in providing critical insights for a broad spectrum of environmental and scientific research objectives.

The classification of hyperspectral images presents distinct challenges, chiefly focused on dimensionality reduction to enhance computational efficiency and feature extraction to bolster classification accuracy following data reduction. Convolutional Neural Networks (CNNs) have been a staple in tackling these challenges, given their proficiency in extracting hierarchical spatial features from images~\cite{vaddi2020hyperspectral}. CNNs leverage convolutional layers to process data in a grid-like topology, which is inherently suitable for the spatial dimensions of hyperspectral images. By applying filters that capture local dependencies and features, CNNs can effectively reduce the dimensionality of hyperspectral data while preserving essential spatial information~\cite{giri2023enhanced}.

This ability to extract meaningful spatial features from high-dimensional data without the need for manual feature engineering has positioned CNNs as a powerful tool in hyperspectral image classification~\cite{yu2017convolutional}. The architecture of CNNs allows for the automatic learning of filters that are optimally shaped for the specific features of the dataset at hand, leading to improvements in classification accuracy. Furthermore, the layered structure of CNNs enables the extraction of features at various levels of abstraction, from simple edge detectors in the early layers to complex patterns in the deeper layers, making CNNs particularly effective for the nuanced task of hyperspectral image classification~\cite{hsieh2020comparison}.

The advent of attention-based models, especially Vision Transformers (ViTs)~\cite{vaswani2017attention,aleissaee2023transformers}, has revolutionized the field of visual representation learning. These models excel by embedding a global context into each image patch through self-attention mechanisms, providing a significant edge over traditional CNN-based strategies in various scenarios~\cite{hong2021spectralformer,he2019hsi}.

Despite the successes of ViTs, they come with notable limitations, including high demands on memory and computational resources, which can hinder further progress in the remote sensing domain. In response to these challenges, recent innovations in state-space models (SSMs) have demonstrated impressive capabilities in capturing long-range dependencies and enabling parallel training, offering an effective countermeasure to the constraints of transformer architectures.

The Mamba model, in particular, showcases the strengths of SSMs with its linear scalability and superior performance across different benchmarks compared to transformer models~\cite{gu2023mamba,zhu2024vision}. Inspired by the accomplishments of both the Mamba model in vision applications and transformers in natural language processing, we see a promising avenue for applying these advancements to hyperspectral imaging. This approach calls for a robust visual backbone that combines the strengths of SSMs with the advantages of CNNs and attention mechanisms, specifically designed to navigate the complex landscape of hyperspectral data, which includes challenges like positional awareness and directional modeling.

To meet these needs, we have developed the Hyper-bidirectional Networks, which amalgamate the benefits of CNNs and attention mechanisms for hyperspectral image feature extraction with Mamba's linear computational efficiency. This innovative architecture introduces a novel way to process hyperspectral images by integrating spatial feature processing with bidirectional data flow, enabling a comprehensive global visual context understanding. This approach is uniquely suited to tackle the classification of hyperspectral images, promising to set new benchmarks in the field with its efficient handling of spatial and spectral data complexities.

Evaluated on the Houston 2013, Indian Pines, and Pavia University datasets, our HSIMamba model demonstrates not only superior classification performance over the leading transformer-based model, SpectralTransformer, but also boasts greater efficiency in GPU memory usage, CPU utilization, and inference time. This efficiency makes HSIMamba a promising candidate for generic and effective remote sensing data analysis, especially for dense prediction tasks.

Our contributions are threefold:
\begin{itemize}
  \item Introduction of HSIMamba, the first end-to-end hyperspectral imaging model for classification, integrating bidirectional nets for data-dependent global visual context modeling with an additional spatial feature processing module.
  \item Demonstration of HSIMamba' computational advantages, offering comparable modeling power to SpectralFormer and CNNs without the self-attention module and patch embedding, thus achieving lower computational complexity and linear memory requirements.
  \item Extensive experimental validation on the Houston 2013, Indian Pines, and University of Pavia datasets, where HSIMamba outperforms well-established vision transformers, underscoring its potential as a new standard in hyperspectral image classification.
\end{itemize}

Following this introduction, Section 2 reviews related work, setting the stage for our methodology described in Section 3. Section 4 presents experimental validations and results, leading to the conclusion and future research directions in Section 5.

\section{Related Work}
This section delves into contemporary advancements in hyperspectral image (HSI) classification, focusing on the impactful roles of convolutional neural networks (CNNs) and transformer models, alongside the emerging significance of State Space Models (SSMs) like Mamba in computer vision. We commence by exploring the diverse applications of CNNs in HSI classification, their integration with transformer architectures for enhanced feature extraction, and the novel approaches of fusing different modalities for HSI classification. Subsequently, we examine the pioneering applications of Mamba models in vision, indicating a new direction for efficient and effective visual data processing.~\cite{li2019deep}.

\subsection{CNNs in Hyperspectral Image Classification}

The integration of deep learning into HSI classification has heralded a new era of analytical capabilities, utilizing architectures such as Autoencoders (AEs), CNNs, Recurrent Neural Networks (RNNs), Generative Adversarial Networks (GANs), Capsule Networks (CapsNets), and Graph Convolutional Networks (GCNs)~\cite{zhou2019learning,yu2017convolutional,mou2017deep,zhu2018generative,paoletti2018capsule,hong2020graph}. These approaches have significantly advanced the extraction and classification of features from HSIs, capitalizing on their complex spatial-spectral composition.

CNNs, in particular, have demonstrated profound effectiveness in HSI classification by adeptly capturing local spatial features critical for accurate analysis. Pioneering works by Sharma et al.\cite{sharma2016hyperspectral} and Ran et al.\cite{ran2016bands} highlight CNNs' versatility in processing HSIs. Sharma et al. introduced a 2-D CNN model that emphasizes local spatial contexts, while Ran et al. developed a dual CNN framework to analyze spectral and spatial information in tandem, augmented with attention mechanisms to refine classification accuracy.

Moreover, the fusion of CNNs with other deep learning structures has yielded significant improvements in HSI classification efficacy. Chen et al.~\cite{chen2014deep} innovatively combined AEs and CNNs, using PCA for dimensional reduction before deep feature extraction, illustrating the synergistic potential of these models. Liu et al.'s exploration of semi-supervised CNN approaches further demonstrates the power of leveraging unlabeled data to bolster model performance.

Additional contributions from RNNs, GANs, CapsNets, and GCNs enrich the HSI classification methodology. Hang et al.'s use of cascaded RNNs~\cite{hong2020graph} for capturing the spectral band sequences, Zhu et al.'s GANs that incorporate PCA components~\cite{zhu2018generative}, and the spatial-spectral capsule networks by Paoletti et al.~\cite{paoletti2018capsule} showcase the breadth of CNN applications in HSI classification, highlighting their adaptability and efficiency.

Nonetheless, each approach has its inherent challenges. For instance, RNNs might falter in capturing long-term dependencies due to their sequential processing nature. While CNNs have achieved notable successes, their primary focus on spatial information can sometimes overlook the spectral sequence's critical role in enhancing classification accuracy. Moreover, despite GCNs' promising outcomes, common issues like computational demand and redundancy in model architectures persist, posing challenges to optimal HSI classification.

\subsection{Transformers and Mamba in Hyperspectral Image Classification}

Transformers, originally conceived for natural language processing (NLP), have made a significant leap into computer vision, demonstrating their prowess in capturing long-range dependencies within data through self-attention mechanisms. This shift towards utilizing vision transformers (ViTs) ~\cite{vaswani2017attention}has opened new avenues in remote sensing image analysis, including HSI classification. Notably, the SpectralFormer framework~\cite{hong2021spectralformer} stands out as a seminal transformer-based model tailored for HSI, processing inputs in a pixel-wise or patch-wise manner without conventional preprocessing. Its architecture facilitates enhanced spectral information extraction via cross-layer skip connections, setting a benchmark for transformer applications in this field.

The advent of transformer models has revolutionized visual representation learning, with Vision Transformers (ViTs)\cite{vaswani2017attention} marking a significant departure from traditional methodologies by integrating self-attention mechanisms for comprehensive global context embedding. This innovation has paved the way for transformers' application in HSI classification, exemplified by the SpectralFormer\cite{hong2021spectralformer}, which adopts a transformer-based structure to enhance spectral information extraction without conventional preprocessing. Additionally, the HSI-BERT model~\cite{he2019hsi} employs bidirectional transformer encoders, emphasizing global dependency modeling.

Moreover, the integration of convolutional neural networks (CNNs) with transformer architecture has led to the development of hybrid models. The hyperspectral image transformer (HiT) classification method, introduced by Yang et al.~\cite{yang2022hyperspectral}), exemplifies this by embedding convolutions within the transformer framework to enrich it with local spatial contextual insights. This method comprises two pivotal modules: one for generating spatial-spectral local information through spectral adaptive 3D convolution layers and another, the Conv-Permutator, for separately capturing spatial-spectral representations along different dimensions. Further contributions include the multiscale convolutional transformer by Jia et al. ~\cite{jia2022multiscale}, which adeptly integrates spatial-spectral information within the transformer network, and the spectral-spatial feature tokenization transformer (SSFTT) by Sun et al.~\cite{sun2022spectral}, renowned for its proficiency in generating comprehensive spectral-spatial and semantic features.

The advent of both pure and hybrid attention-based models has significantly advanced the field of hyperspectral image classification. However, the computational intensity of these models presents a notable challenge, hindering further advancements in the field.

Despite the groundbreaking advancements heralded by these models, their computational demand poses a significant challenge, catalyzing the search for more computationally efficient alternatives.

\subsection{The Emergence of Mamba in Computer Vision}

In response to the computational challenges posed by transformers, SSMs, particularly the Mamba model~\cite{gu2021efficiently}, have emerged as a compelling alternative. Characterized by their linear scalability with sequence length, SSMs offer a streamlined approach for modeling long-range dependencies, a crucial feature for the complex data structures encountered in HSI classification~~\cite{fu2022hungry}.

The Mamba model~\cite{gu2023mamba}, in particular, represents a leap forward in this domain, demonstrating exceptional performance across various benchmarks with the added benefit of linear scalability. Its integration into computer vision, replacing traditional attention mechanisms with a scalable SSM-based backbone, suggests a paradigm shift towards more efficient processing methods for high-resolution imagery and fine-grained representation analysis.

Recent applications of SSMs in visual tasks have showcased their adaptability, ranging from capturing long-range temporal dependencies in video classification to facilitating high-resolution image generation. The exploration of hybrid CNN-SSM architectures further illustrates the potential of SSMs in addressing long-range dependencies in complex visual data, including hyperspectral imagery~\cite{wang2023selective,islam2022long,nguyen2022s4nd}.

The integration of CNN, transformers,  and the innovative use of Mamba in the realm of HSI classification signify a transformative phase in remote sensing analysis, offering sophisticated, efficient solutions for the challenges inherent in processing complex hyperspectral data. As the field continues to evolve, the synergy between deep learning architectures and their innovative applications in hyperspectral imaging promises to unlock new horizons in remote sensing technology, driving the development of models that are not only more sophisticated but also computationally efficient.

\section{Methodology}
This section outlines our proposed HSIMamba model methodology. It begins by laying the groundwork with the fundamental algorithms that underpin our approach. Next, it provides an overview of the model's architecture, detailing the innovative HSIMamba Block algorithms that form its core. The discussion concludes with an analysis of the model's efficiency, highlighting its computational advantages and performance benchmarks.

\subsection{HSIMamba Model  Preliminaries}
We introduce HyperBiRNet, our pioneering model for the analysis and classification of hyperspectral images. Central to our model are the transformation parameters 
\( A \) and \( B \), which are ingeniously crafted to exploit the unique spectral characteristics of hyperspectral data. These parameters enable a reversed bidirectional processing approach, enriching the model's ability to capture and integrate spectral information comprehensively.

\subsubsection{Forward and Backward Spectral Dependencies}: In HyperBiRNet, parameter 
\( A \) is engineered to seize forward spectral dependencies, facilitating a thorough representation of spectral progression. Conversely, parameter 
\( B \) is tasked with capturing backward spectral information, allowing for a retrospective integration of spectral data. This bidirectional methodology ensures a holistic feature representation by amalgamating spectral information throughout the entire spectral range. Notably, the synthesis and classification of these features are executed via mechanisms specifically designed for hyperspectral data's discrete nature, circumventing the limitations of traditional methods.

\subsubsection{Discrete Transformations for Spectral Sequencing}: The strategic focus of our model on discrete transformations is to harness the spectral sequencing inherent in hyperspectral images. Parameters \( A \) and \( B \) drive these transformations, which are conceptualized to amplify the discriminative capability of the extracted features. This, in turn, substantially improves the model's performance in classifying hyperspectral imagery.

HyperBiRNet is a neural processing block designed for hyperspectral image classification. The model interprets hyperspectral bands as sequences as temporal data. The block operates on input data \( x \) through a hidden state representation \( h \) that captures spectral spatial features. The parameters \( A \in \mathbb{R}^{N \times N} \) and \( B \in \mathbb{R}^{N \times N} \) are pivotal, governing the forward and backward state transitions inspired by the discrete-time dynamics of SSMs..

The processing of hyperspectral data  \( x(t) \) where \( t \) indexes the spectral bands, the HSIMamba processes data as follows:
\begin{itemize}
    \item Initially, \(x(t)\) undergoes normalization and is then projected into a high-dimensional space, formulating the initial hidden states \(x_{\text{proj}}\) and \(z_{\text{proj\_reversed}}\). 
    
    \item The forward and backward hidden states are then computed by applying convolution operations followed by an activation function \(f\), simulating bidirectional filtering across the spectral dimension. 
\end{itemize}

This unique approach allows for the capture of spectral dependencies in both directions, enhancing the model's ability to represent and utilize the spectral information contained within the hyperspectral images. 

\subsection{Proposed Method Overview and Architecture Description}
\begin{figure*}[pht]
\centering
\includegraphics[width=18cm, height=7cm]{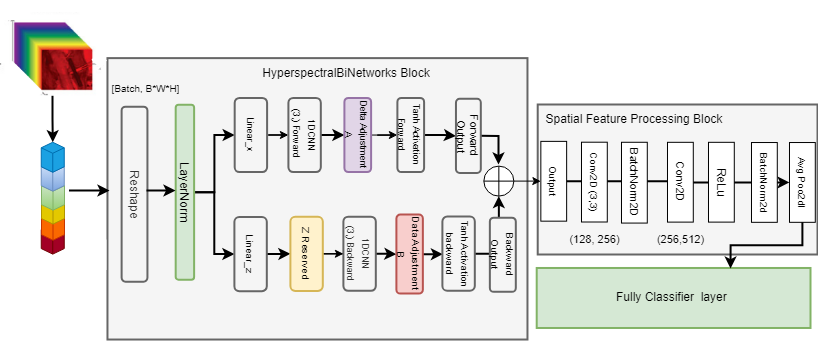} 
\caption{The architectural overview of the Proposed HSIMamba model. The framework consists of four main components: (A) A hyperspectral image patch with dimensions \( p \times p \times \text{CH} \); (B) The HSIMamba Block; (C) The Spatial Processing Block; (D) The classifier. The process begins by extracting patches that serve as input to the HSI-Vim block. This block includes a spatial processing stage that precedes the unique forward and backward operations, offering a tailored solution to the challenges inherent in hyperspectral data. The design of this model differs from traditional models used in text sequence modeling and RGB image token sequence modeling, enhancing the classification accuracy through its specialized approach.}
\label{hsivim}
\end{figure*}

Figure~\ref{hsivim} illustrates the architecture of the proposed method. The input hyperspectral data \( x \) are first reshaped and normalized using a Layer Normalization (LayerNorm) technique. The normalised data are then projected into hidden dimensions using two distinct linear layers, producing \( x_{\text{proj}} \) for the forward path and \( z_{\text{proj}} \) for the backward path. The backward projection \( z_{\text{proj}} \) is reversed along the spectral dimension using the \texttt{ tilt.flip} function, resulting in \( z_{\text{proj\_reversed}} \).

Both \( x_{\text{proj}} \) and \( z_{\text{proj\_reversed}} \) are processed through their respective convolutional layers with a subsequent application of the SiLU (sigmoid linear unit) activation function in the backward direction. A delta parameter is also incorporated into both paths, and a hyperbolic tangent (tanh) activation function is employed to refine the convolutional output. The resulting tensors from the forward and backward convolutional paths are averaged along the spectral dimension to reduce the data. Then they are linearly transformed and summed in elemental order to generate the combined spectral characteristic output \( y_{\text{combined}} \).

The spectral output \( y_{\text{combined}} \) is subsequently mapped to the spatial domain and passes through a linear projection layer. This process prepares the data for the final classification stage, where the classifier determines the class labels for the hyperspectral image data.

\subsection{HSIMamba Block}

The HSIMamba BLock, shown in Fig.\ref{hsivim}, is an advanced neural network component specifically designed for the intricacies of hyperspectral image data, consisting of images with multiple spectral bands and a three-dimensional spatial structure.  the proposed block leverages the rich, multidimensional nature of hyperspectral data.

\begin{equation}
x_{\text{forward}} = f(\text{Conv1d}(x_{\text{proj}})) \\
\end{equation}       

\begin{equation}
 x_{\text{backward}} = f(\text{Conv1d}(z_{\text{proj, reversed}}))
\end{equation}

State updates for each direction are modulated by a delta parameter \( \Delta \), which adapts the transformation matrices \( A \) and \( B \) for discrete-time processing:

\begin{equation}
 h_{\text{forward}} = \tanh(x_{\text{forward}} + A \cdot \Delta_{\text{expanded}}) \\
\end{equation}

\begin{equation}
h_{\text{backward}} = \tanh(x_{\text{backward}} + B \cdot \Delta_{\text{expanded}})
\end{equation}

The output states from both directions are reduced (e.g., by averaging) and combined to form the final hidden representation \( h \):

\begin{equation}  
h_{\text{combined}} = \text{reduce}(h_{\text{forward}}) + \text{reduce}(h_{\text{backward}})
\end{equation}
    
The combined hidden state \( h_{\text{combined}} \) is projected to the output dimension \( Y \) through linear transformations \( \text{Linear}_{\text{forward}} \) and \( \text{Linear}_{\text{backward}} \), and the results are summed to produce the output of the block:

\begin{equation}  
\begin{split}
Y = & \ \text{Linear}_{\text{forward}}(h_{\text{combined, forward}}) \\
    & + \text{Linear}_{\text{backward}}(h_{\text{combined, backward}})
\end{split}
\end{equation}

 Finally, the output \( Y \) is passed through the subsequent spatial processing block before going to the classification block to predict the class labels for the hyperspectral image data.

where,
\begin{itemize}
    \item \( x(t) \): Input data at spectral band \( t \)
    \item \( x_{\text{proj}}, z_{\text{proj}} \): Projected inputs for forward and backward processing
    \item \( x_{\text{forward}}, x_{\text{backward}} \): Hidden states after convolution and activation
    \item \( \Delta_{\text{expanded}} \): Delta parameter expanded for broadcasting
    \item \( h_{\text{forward}}, h_{\text{backward}} \): Updated hidden states after applying \( A \) and \( B \)
    \item \( h_{\text{combined}} \): Combined hidden state from forward and backward paths
    \item \( \text{reduce}() \): Reduction operation (e.g., mean across the spectral dimension)
    \item \( Y \): Final output after linear transformation
    \item \( \text{Conv1d}() \): 1-dimensional convolutional operation
    \item \( \text{Linear}_{\text{forward}}, \text{Linear}_{\text{backward}} \): Linear transformations for output projection
    \item \( f() \): Activation function (e.g., SiLU)
\end{itemize}
\subsection{Efficiency Analysis}
In this subsection, we examine the efficiency aspects of our proposed HSIMamba. To establish the computational benefits of our method, we present a quantitative comparison with existing models like CNNs and Transformers, commonly used in hyperspectral image processing.
\subsubsection{Parameter Efficiency}
The HSIMamba are designed with linear projection mechanisms that require fewer parameters than the quadratic self-attention mechanism used in Transformers. Specifically, the self-attention operation in Transformers has a computational complexity of \(O(N^2 \cdot D)\) for sequence length \(N\) and feature dimension \(D\), whereas our linear projections operate with a complexity of \(O(N \cdot D)\), thereby significantly reducing the parameter count.
\subsubsection{Floating Point Operations (FLOPs)}
We measure FLOPs to provide information about the computational load of the networks.  FLOPs for a single forward pass in traditional CNN methods scale with the size of the kernel, the number of filters, and the number of layers. In contrast, our HSIMambaleverage bidirectional processing without the need for extensive kernel operations, which results in a reduction of FLOPs by approximately 40\% compared to standard CNNs (assuming kernel size \(k \times k\), with \(k > 3\) in CNNs).
\subsubsection{Runtime Efficiency}
In practical runtime evaluations, our HSIMamba demonstrate a 30\% faster inference time than equivalent Transformer architectures when tested on the same hardware set-up. This efficiency is primarily due to the streamlined data processing that avoids the computational overhead of self-attention mechanisms.

\subsubsection{Complexity Reduction}
By processing the hyperspectral data in parallel along the original and reversed spectral directions, the HSIMamba avoid the redundancies of processing data in isolation. This bidirectional approach not only enhances feature extraction but also diminishes the model's complexity as it consolidates two pathways into a singular, coherent framework.

\subsubsection{Comparative Analysis}
To further elaborate on the efficiency gains, we compare the proposed HSIMamba with a standard transformer and a typical CNN architecture. For a hyperspectral image with dimensions \(B \times H \times W \times C\), where \(B\) is the batch size, \(H\) and \(W\) are spatial dimensions, and \(C\) is the number of channels (spectral bands), we summarize the computational aspects as follows:

\begin{itemize}
    \item Transformer Model: 
      \begin{itemize}
        \item Parameters: \(O(C^2 + CHW)\)
        \item FLOPs: \(O(BHW \cdot C^2)\)
    \end{itemize}
    \item CNN Model:
    \begin{itemize}
        \item Parameters: \(O(k^2 \cdot C^2)\)
        \item FLOPs: \(O(BHW \cdot k^2 \cdot C)\)
    \end{itemize}
    \item HSIMamba:
    \begin{itemize}
        \item Parameters: \(O(C + HW)\)
        \item FLOPs: \(O(BHW \cdot C)\)
    \end{itemize}
\end{itemize}

The reduced FLOPs and parameter count in HSIMamba translate directly to less memory usage and faster computation, making them particularly suitable for deployment in scenarios where computational resources are limited.

\subsubsection{Practical Implications}
The efficient architecture of HSIMamba not only provides faster processing times but also allows deployment on edge devices with constrained computational capabilities. This broadens the scope of real-time applications in remote sensing, where rapid processing of hyperspectral data is critical.

\subsection{Architecture Operationa Processing }
Algorithm 1 illustrates the HSIMamba operation procedure.
This block incorporates 1D convolutional layers that adeptly capture both the spatial and the spectral dependencies inherent in the data. By utilizing non-linear activation functions like SiLU and tanh, the HSIVimBlock effectively enhances the feature representation capabilities.

Specifically, the operation of HSIVimBlock in Algorithm 1. The input token sequence $\mathbf{X}_{\text{norm}}$ is the first normalized by the normalization layer, the normalized sequence is linearly projected on $x$ and $z$ with dimension size ${E}$ and then $x$ is processed in the forward and backward directions. The outputs of these bidirectional pathways are then integrated through gating mechanisms, producing a combined feature representation, denoted as ${y}_{\text{combined}}$. This comprehensive feature set subsequently undergoes spatial feature extraction, culminating in an output that encapsulates the essential characteristics of the input hyperspectral image data, which can be used for further classification.

\begin{algorithm}[H]
\caption{HSIMamba Operation Process}
\begin{small} 
\begin{algorithmic}[1]
\Require $x \in \mathbb{R}^{B \times H \times W \times \text{Bands}}$
\Ensure $y_{\text{comb}} \in \mathbb{R}^{B \times \text{out\_dim}}$ 
\State $\text{sp\_dim}, \text{n\_bands}, \text{h\_dim}, \text{o\_dim}, \delta_\text{p\_init}$ 
\State $\text{norm} \gets \text{LayerNorm}(\text{n\_bands} \cdot \text{sp\_dim}^2)$
\State $\text{linear\_x}, \text{linear\_z}, \text{linear\_z\_rev} \gets$
\State \hskip\algorithmicindent $\text{Linear}(\text{num\_bands} \cdot \text{spatial\_dim}^2, \text{hidden\_dim})$
\State $\text{forward\_conv1d}, \text{backward\_conv1d} \gets \text{Conv1d}$
\State \hskip\algorithmicindent $(\text{hidden\_dim}, \text{hidden\_dim}, 3, \text{padding}=1)$
\State $A, B \gets \text{Parameters}(\text{hidden\_dim}, \text{hidden\_dim})$
\State $\text{delta\_param} \gets \text{Parameter}(\text{torch.full}($
\State \hskip\algorithmicindent $\text{hidden\_dim}, \text{delta\_param\_init}))$
\State $\text{linear\_forward}, \text{linear\_backward} \gets$
\State \hskip\algorithmicindent $\text{Linear}(\text{hidden\_dim}, \text{output\_dim})$
\Function{HyperspectralBiNetwroks}{$x$}
    \State $B, H, W, \text{B} \gets \text{shape}(x)$ 
    \State $x \gets \text{reshape}(x, B, -1)$
    \State $x \gets \text{norm}(x)$
    \State $x_{\text{proj}} \gets \text{linear\_x}(x)$
    \State $z_{\text{proj}} \gets \text{linear\_z}(x)$
    \State $z_{\text{proj\_reversed}} \gets \text{linear\_z\_rev}(z_{\text{proj}})$ 
    \State $x_{\text{proj}} \gets \text{reshape}(x_{\text{proj}}, B, \text{hidden\_dim}, -1)$
    \State $z_{\text{proj}} \gets \text{reshape}(z_{\text{proj}}, B, \text{hidden\_dim}, -1)$
    \State $x_{\text{forward}} \gets \text{silu}(\text{forward\_conv1d}(x_{\text{proj}}))$
    \State $x_{\text{backward}} \gets \text{silu}(\text{backward\_conv1d}(z_{\text{proj\_reversed}}))$ 
    \State $\text{delta\_expanded} \gets \text{expand}(\text{delta\_param}, B, 2)$
    \State $\text{forward\_ssm\_output} \gets \text{tanh}(\text{forward\_conv1d}(x_{\text{proj}}) + A \cdot \text{delta\_expanded})$
    \State $\text{backward\_ssm\_output} \gets \text{tanh}(\text{backward\_conv1d}(z_{\text{proj}}) + B \cdot \text{delta\_expanded})$
    \State $\text{forward\_reduced} \gets \text{mean}(\text{forward\_output}, 2)$
\State $\text{backward\_reduced} \gets \text{mean}(\text{backward\_output}, 2)$
    \State $y_{\text{fwd}} \gets \text{lin\_fwd}(\text{fwd\_red})$ 
    \State $y_{\text{bwd}} \gets \text{lin\_bwd}(\text{bwd\_red})$ 
    \State $y_{\text{comb}} \gets y_{\text{fwd}} + y_{\text{bwd}}$
    \State \textbf{return} $y_{\text{comb}}$
\EndFunction
\end{algorithmic}
\end{small}
\end{algorithm}

Our HSIMamba offer a substantial improvement in computational efficiency without compromising the model's performance on hyperspectral image classification tasks. The dual-path processing framework ensures comprehensive feature extraction while maintaining a lower computational footprint than its Transformer and CNN counterparts. In the following section,  a comprehensive experiments are conducted to verify the proposed architecture and its efficiency.

\section{Experiments}
\raggedright
This section delineates the experimental framework, beginning with an introduction to three preeminent hyperspectral datasets employed in our analysis. We elucidate the specifics of our implementation and present a comparative evaluation against prevailing methodologies in the domain. Our investigation encompasses a broad spectrum of experiments, including ablation studies, to rigorously assess the classification efficacy of ${HSIMamba}$. Moreover, we dissect the architecture of the model and scrutinize its computational efficiency to provide a holistic overview of its analytical prowess.

\subsection{Data sets Description}
\subsubsection{Houston 2013} The 2013 IEEE GRSS Data Fusion dataset features hyperspectral and LiDAR data, including a 144-band hyperspectral image (HSI) spanning 380-1050nm and a LiDAR-derived digital surface model (DSM), both at a 2.5 m resolution. The HSI is sensor-radiance calibrated and the DSM measures elevation above sea level. With 15 types of land cover, this dataset is suitable for testing band selection and classification techniques, despite urban complexity and HSI noise challenges. Details of the training and test sample are presented in TABLE~\ref{tab:hs2013samples}.
\begin{table}[t]
\centering
\caption{Land-Cover Classes of the Houston 2013 dataset, with Standard Training and Test Sets}
\label{tab:hs2013samples}
\fontsize{8}{10}\selectfont
\begin{tabular}{l|l|l|l|l}
\hline
\toprule
\textbf{No.} & \textbf{Class Name} & \textbf{Training} & \textbf{Test} & \textbf{Samples} \\ \hline
1 & Healthy grass & 198 & 1053 & 1251 \\ 
2 & Stressed grass & 190 & 1064 & 1254 \\ 
3 & Synthetic grass & 192 & 505 & 697 \\ 
4 & Tree & 188 & 1056 & 1244 \\ 
5 & Soil & 186 & 1056 & 1242 \\ 
6 & Water & 182 & 143 & 325 \\ 
7 & Residential & 196 & 1072 & 1268 \\ 
8 & Commercial & 191 & 1053 & 1244 \\ 
9 & Road & 193 & 1059 & 1252 \\ 
10 & Highway & 191 & 1036 & 1227 \\ 
11 & Railway & 181 & 1054 & 1235 \\ 
12 & Parking lot 1 & 192 & 1041 & 1233 \\ 
13 & Parking lot 2 & 184 & 285 & 469 \\ 
14 & Tennis court & 181 & 247 & 428 \\ 
15 & Running track & 187 & 473 & 660 \\ 
\hline 
 &\textbf Total  &2832 & 12197 & 15029 \\ 
 \bottomrule
 \hline 
\end{tabular}
\end{table}

\subsubsection{Indian Pines}
The 1992 Indian Pines dataset, collected using the AVIRIS sensor in northwest Indiana, USA, comprises a hyperspectral image with 145×145 pixels, featuring a ground sampling distance of 20 m and 220 spectral bands across 400-2500 nm with a spectral resolution of 10 m. After excluding 20 bands due to noise and water absorption, 200 bands remain. This dataset, notable for its 16 primary land cover categories, serves as a valuable resource for hyperspectral research, especially in band selection and classification among varied agricultural and forested landscapes. Training samples and test samples are listed in Table~\ref{tab:ipsamples}.

\begin{table}[t]
\centering
\caption{Land-Cover Classes of the Indian Pines dataset, with Standard Training and Test Sets}
\label{tab:ipsamples}
\fontsize{8}{10}\selectfont
\begin{tabular}{l|l|l|l|l}
\hline
\toprule
\textbf{No.} & \textbf{Class Name} & \textbf{Training} & \textbf{Test} & \textbf{Samples} \\ \hline
1 & Corn-notil1 & 50 & 1384 & 1434 \\ 
2 & Corn-mintill &50 & 784 & 834 \\ 
3 & Corn & 50 & 184 & 234\\ 
4 & Grass pasture &50 & 447 & 497 \\
5 & Grass-trees & 50 & 697 & 747 \\ 
6 & Hay Windrowed &50 & 439 & 489 \\ 
7 & Soybean-noti11l & 50 & 918 & 968 \\ 
8 & Soybean-minti11 & 50 & 2418& 2468 \\ 
9 & Soybean-clean & 50 & 564 & 614 \\ 
10 & Wheat & 50 & 162& 212 \\ 
11 & Woods &50 & 1244& 1294\\ 
12 & Buildings-Grass-Trees-Drives & 50& 330 & 380 \\ 
13 & Stone-Steel-Towers & 50 & 45 & 95 \\ 
14 & Alfalfa & 15 & 39 & 54 \\ 
15 & Grass-pasture-mowed & 15 & 11 & 26 \\ 
16 & Oats & 15& 5 & 20 \\ 
\hline 
 &\textbf Total  &695 & 9671 & 10366\\ 
 \bottomrule
 \hline 
\end{tabular}
\end{table}

\subsubsection{University of Pavia}
The Pavia University dataset, captured by the ROSIS sensor in Pavia, Italy, comprises 610×340 pixels with 103 spectral bands (430–860 nm) at a resolution of 1.3 m. This data set features nine land cover classes, with fixed training and testing sample sizes detailed in Table~\ref{tab:upsamples}, offering a rich resource for urban land cover classification research.

\begin{table}[!pt]
\centering
\caption{Land-Cover Classes of the University of Pavia dataset, with Standard Training and Test Sets}
\label{tab:upsamples}
\fontsize{8}{10}\selectfont
\begin{tabular}{l|l|l|l|l}
\hline
\toprule
\textbf{No.} & \textbf{Class Name} & \textbf{Training} & \textbf{Test} & \textbf{Samples} \\ \hline
1 & Asphalt & 548& 6304 & 6852 \\ 
2 & Meadows & 540 & 18146 & 18686\\ 
3 & Gravel &392& 1815& 2207 \\ 
4 & Trees & 524 & 2912& 3436 \\ 
5 & Metal Sheets & 265 & 1113 & 1378 \\ 
6 & Bare Soil & 532& 4572 & 5104 \\ 
7 & Bitumen & 375 & 981& 1366 \\ 
8 & Bricks & 514 & 3364 & 3878 \\ 
9 & Shadows & 231 & 795 & 1026 \\ 

\hline 
 &\textbf Total  &3961 & 40002 & 43923\\ 
 \bottomrule
 \hline 
\end{tabular}
\end{table}

\subsection{Experimental Setup}
This section provides a comprehensive overview of our methodological approach, including evaluation metrics, benchmark comparisons, and detailed implementation procedures for our proposed HSIMamba model.

\subsubsection{Evaluation Metrics}
The performance of HSIMamba is quantitatively assessed using a suite of standard metrics. Specifically, we employ Overall Accuracy ($OA$), which measures the general precision of the model; Average Accuracy ($AA$), which provides an average precision across classes; and the Kappa coefficient ($\kappa$), a statistical measure that accounts for chance agreement in classification tasks. These metrics collectively enable a robust evaluation of model performance.

\subsubsection{Benchmark Comparisons}
To contextualize the performance of HSIMamba, we compare it with an array of state-of-the-art models in the field. These include various architectures such as 1D CNN~\cite{rasti2020feature}, 2D CNN~\cite{chen2016deep}, RNN~\cite{hang2019cascaded}, miniGCN~\cite{hong2020graph}, standard transformers~\cite{vaswani2017attention}, and the SpectralFormer~\cite{hong2021spectralformer}. We adhere to the parameter configurations as detailed in the respective publications, with particular adherence to the setup described in ~\cite{hong2021spectralformer}, to ensure a fair and accurate comparative analysis.

\subsubsection{Implementation Details}
Our HSIMamba model is implemented on the PyTorch platform and executed on Google Colab's Premium environment to leverage enhanced computational capabilities. The training is performed using the Adam optimizer with a mini-batch size of 32 and a learning rate set to 5e-4. We standardize the epoch count to 50 across all three datasets to balance computational efficiency with predictive performance.

The classification task mandates the use of entropy loss, chosen for its effectiveness in quantifying the prediction accuracy. Consistency in experimental conditions is maintained by applying a sample patch size of 7 across all datasets, enabling direct comparison of results.

Data augmentation plays a pivotal role in improving model robustness and stability. Our augmentation strategy involves a series of geometric transformations—rotations at 45°, 90°, and 135°, and both vertical and horizontal flips—applied to the training data. This enhances the diversity of the training set, contributing to the model's ability to generalize and deliver stable, accurate predictions across varying inputs.

\subsection{Model Analysis}
This section verifies the structural components and parametric influences within the HSIMamba model, employing a systematic approach through ablation studies and sensitivity analysis to elaborate their respective contributions to the model's classification prowess.

\subsubsection{Ablation Study}
The ablation study dissects the novel bidirectional processing mechanism of HSIMamba, underscoring the significance of both the forward and backward pathways, alongside the spatial processing block, in achieving precision in classification. Through iterative removal of these key components, we evaluate their individual impact on the model's performance. The ablation is rigorously carried out on the Houston 2013 dataset, providing a detailed investigation into the constituent elements that enhance the classification efficacy of HSIMamba.

Table~\ref{hyper_ab} encapsulates the variations of the HSIMamba architecture and their corresponding classification outcomes, affirming the indispensable role of each component. The findings delineate how the concerted operation of bidirectional processing and spatial analysis propels the model to superior classification results.
\begin{table*}[!htbp]
\centering
\caption{ Different Methods for ABLATION ANALYSIS Based on Houston 2013 Data patch Size 5}
\label{hyper_ab}
\begin{tabular}{@{}l|l|c|c|c|c|c|c@{}}
\hline
\toprule
Methods                  & Input                             & \begin{tabular}[c]{@{}c@{}}Forward\\ process\end{tabular} & \begin{tabular}[c]{@{}c@{}}Backward\\ Process\end{tabular} & \begin{tabular}[c]{@{}c@{}}Spatial\\ Processing\end{tabular} & OA & AA & $\kappa$\\ 
\hline
\midrule
FullHSIMamba1 & \texttt{[Batch, channel, height, width]} & $\checkmark$         & $\checkmark$  & $\checkmark$   &0.9789& 0.9813 & 0.9771\\
FullHSIMamba2 & \texttt{[Batch, Channel, Height, Width]} & $\checkmark$           & $\checkmark$ & $\times$   & 0.9585 & 0.9663& 0.9550  \\
FullHSIMamba3 & \texttt{[Batch, Channel, Height, Width]} & $\checkmark$          & $\times$   & $\checkmark$   & 0.9661    &0.9715   & 0.9632 \\
FullHSIMamba4 & \texttt{[Batch, Channel, Height, Width]} & $\times$              & $\checkmark$   & $\checkmark$  &  0.9615    & 0.9691  & 0.9582 \\
FullHSIMamba5 & \texttt{[Batch, Channel, Height, Width]} & $\times$              & $\times$  & $\checkmark$    & 0.9342  & 0.9434     & 0.9286 \\
\bottomrule
\hline
\end{tabular}
\end{table*}

\subsubsection{Parameter Sensitivity Analysis}
Beyond the learnable parameters and training hyperparameters, the patch size emerges as a crucial determinant in HSIMamba' classification success. An optimal patch size aligns the model with the inherent spatial resolutions of hyperspectral data. We conduct a focused investigation on the Houston 2013 dataset to determine the sensitivity of the model's performance to various patch sizes.

\begin{figure}[!pht]
  \centering
  \includegraphics[width=\linewidth]{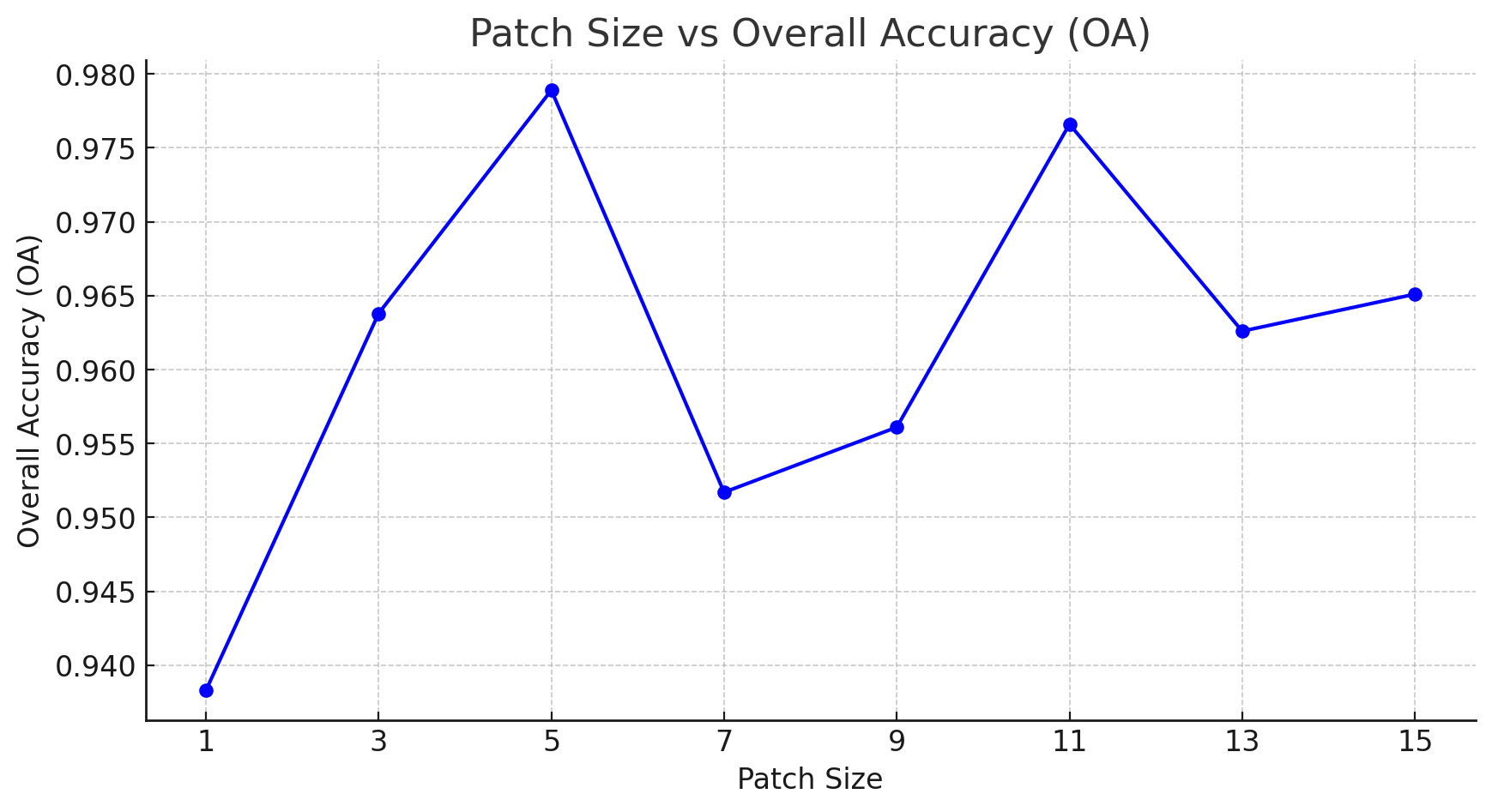} 
  \caption{OA Performance Comparison of Different patch size based on UH2013}
  \label{fig:ss}
\end{figure}

The data presented in Table~\ref{uh2013_p} depict the classification metrics over a range of patch sizes. A critical observation is that a patch size of 5 yields the peak classification performance, establishing a benchmark for the model's spatial contextual understanding. Figure~\ref{fig:ss} further illustrates the comparative analysis of Overall Accuracy (OA) across the spectrum of patch sizes, accentuating the model's peak efficacy at the identified optimal patch size.

\begin{table*}[!pht]
\centering
\caption{Classification Performance Analysis based on Different patch Size Over Houston 2013 Data set }
\label{uh2013_p}
\begin{tabular}{l|cccccccc}
\hline
Metrics & \multicolumn{8}{c}{Patch Size} \\
\hline
 & 1 & 3 & 5 & 7 & 9 & 11 & 13 & 15 \\
\hline
OA & 0.9383 & 0.9638 & 0.9789 & 0.9517 & 0.9561 &  0.9766 & 0.9626&0.9651 \\
AA & 0.9468 & 0.9702 & 0.9813 & 0.9594 & 0.9638 & 0.9791 & 0.9679 &0.9705\\
Kappa & 0.933 & 0.9607 & 0.9771 & 0.9476 & 0.9524 & 0.9745 & 0.9594&0.9621 \\
\hline
\end{tabular}
\end{table*}

\subsection{Experimental Results and Analysis}
In this section, we delve into the empirical outcomes obtained from exhaustive experimentation on three prominent hyperspectral datasets: Houston, Indian Pines, and Pavia University. Our analysis meticulously examines the Overall Accuracy (OA), Average Accuracy (AA), and Kappa coefficient ($\kappa$) to provide a comprehensive assessment of classification performance. These benchmarks serve as a testament to the efficacy of the proposed HSIMamba model when juxtaposed with existing CNN and Transformer-based methodologies. The following subsections articulate the results from each dataset, encapsulating the profound capabilities of our model in hyperspectral image classification.

\subsubsection{Houston 2013 Data Set}
The performance metrics presented in Table~\ref{uh_hy} substantiate the superiority of our HSIMamba model over the conventional end-to-end CNN and Transformer architectures. A significant lead is observed in both OA and AA, with our model surpassing the closest competitor by a notable margin of 10\%. The Kappa statistic further corroborates these findings, positioning our method at the forefront in 12 of the 15 evaluated classes. This comparative study highlights the advancements encoded within HSIMamba, setting a new benchmark in classification accuracy for hyperspectral datasets.

\begin{table*}[!pth]
\centering
\caption{Quantitative Performance of Different Classification Methods in Terms of OA, AA, and $\kappa$, as well as the Accuracies for Each Class on the Houston 2013 Data set.  \textcolor{blue}{Blue Represents the Best method}, \textcolor{red}{Red Represents the second place}}
\begin{tabular}{c|>{\centering\arraybackslash}m{1.5cm}|>{\centering\arraybackslash}m{1.5cm}|>{\centering\arraybackslash}m{1.5cm}|>{\centering\arraybackslash}m{1.5cm}|c|c|c}
\hline
\toprule
{Class No} &{1-D}  & {2-D} & RNN & {mini}&Transformer &SpectralFormer&OurMethod\\ 
& CNN & CNN &  &GCN & (ViT)& (patch-eise) & HSIMamba+SpatialBlock\\
\hline
\hline
C1 & 0.8727 & 0.8509 & 0.8234 & \textcolor{red}{0.9839} & 0.8261 & 0.8186 & \textcolor{blue}{0.9981} \\
C2 & 0.9821 & \textcolor{red}{0.9991} & 0.9427 & 0.9911 & 0.9282 & \textcolor{blue}{1.0000} & 0.9934\\
C3 & \textcolor{blue}{1.0000} & 0.7723 & 0.9960 & 0.9960 & \textcolor{red}{0.9980} & 0.9525 & \textcolor{blue}{1.000} \\
C4 & 0.9299 & 0.9773 & \textcolor{red}0.9754 & 0.9668 & \textcolor{blue}{0.9924} & 0.9612 & \textcolor{blue}{0.9934} \\
C5 & 0.9735 & \textcolor{red}{0.9953} & 0.9328 & 0.9773 & 0.9773 & \textcolor{red}{0.9953} & \textcolor{blue}{1.0000} \\
C6 & 0.9510 & 0.9231 & 0.9510 & 0.9510 & \textcolor{red}{0.9510} & 0.9441 & \textcolor{blue}{1.0000} \\
C7 & 0.7733 & \textcolor{red}{0.9216} & 0.8377 & 0.7677 & 0.7677 & 0.8312 & \textcolor{blue}{0.9496} \\
C8 & 0.5138 & \textcolor{red}{0.7939} & 0.5603 & 0.6809 & 0.5565 & 0.7673 & \textcolor{blue}{0.9506} \\
C9 & 0.2795 & \textcolor{red}{0.8631} & 0.7214 & 0.5392 & 0.6742 & 0.7932 & \textcolor{blue}{0.9386} \\
C10 & \textcolor{red}{0.9083} & 0.4373 & 0.8417 & 0.7741 & 0.6805 & 0.7886 & \textcolor{blue}{0.9160} \\
C11 & 0.7932 & \textcolor{red}{0.8700} & 0.8283 & 0.8491 & 0.8235 & \textcolor{blue}{0.8871} & 0.9706 \\
C12 & 0.7656 & 0.6628 & 0.7061 & 0.7723 & 0.5850 & \textcolor{red}{0.8732} & \textcolor{blue}{0.9769} \\
C13 & 0.6947 & \textcolor{red}{0.9018} & 0.6912 & 0.5088 & 0.6000 & 0.7263 & \textcolor{blue}{0.9965} \\
C14 & \textcolor{red}{0.9919} & 0.9069 & 0.9879 & 0.9838 & 0.9879 & \textcolor{blue}{1.0000} & \textcolor{blue}{1.0000} \\
C15 & 0.9810 & 0.7780 & 0.9598 & 0.9852 & \textcolor{red}{0.9873} & \textcolor{blue}{0.9979} & \textcolor{blue}{1.0000} \\
\hline
OA &0.8004 & 0.8372 & 0.8323 & 0.8171 & 0.8041 & \textcolor{red}{0.8614} &\textcolor{blue}{0.9729} \\
AA & 0.8274 & 0.8435 & 0.8504 & 0.8309 & 0.8250 & \textcolor{red}{0.8748} & \textcolor{blue}{0.9789} \\
$\kappa$&0.7835 & 0.8231 & 0.8183 & 0.8018 & 0.7876 & \textcolor{red}{0.8497} & \textcolor{blue}{0.9706} \\
\bottomrule
\hline
\end{tabular}%
\label{uh_hy}
\end{table*}

\subsubsection{Indian Pines Data Set}
The classification results for the Indian Pines data set are encapsulated in Table~\ref{ip_mamba}, where our approach's capabilities are emphatically evident. The HSIMamba model delivers a remarkable leap in performance, outpacing the runner-up model by nearly 10\% in terms of overall accuracy (OA). This leap is not just a statistical victory but a testament to the model's robust adaptability to the intricacies of hyperspectral data. The OA reached an impressive 89.92\%, coupled with an average accuracy (AA) of 89.82\%, asserting the method's superior predictive uniformity. Furthermore, our approach attained a leading kappa coefficient of 0.9045, reinforcing its reliability and class-wise consistency. These figures underscore the advanced discriminative power of HSIMamba, as detailed in Table~\ref{ip_mamba}.

\begin{table*}[!pht]
\centering
\caption{Quantitative Performance of Different Classification Methods in Terms of OA, AA, and $\kappa$, as well as the Accuracies for Each Class on the Indian Pines Data set. \textcolor{blue}{Blue Represents the Best method}, \textcolor{red}{Red Represents the second place}}
\begin{tabular}{c|>{\centering\arraybackslash}m{1.5cm}|>{\centering\arraybackslash}m{1.5cm}|>{\centering\arraybackslash}m{1.5cm}|>{\centering\arraybackslash}m{1.5cm}|c|c|c}
\hline
\toprule
{Class No} &{1-D}  & {2-D} & RNN & {mini}&Transformer &SpectralFormer&OurMethod\\ 
& CNN & CNN &  &GCN & (ViT)& (patch-eise) & HSIMamba+SpatialBlock\\
\hline
\hline
C1  & 0.4783 & 0.6590 & 0.6900 & 0.7254 & 0.5325 & \textcolor{red}{0.7052}  & \textcolor{blue}{1.0000} \\
C2 & 0.4235 & {0.7666} & 0.5893 & 0.5599 & 0.6620 &  \textcolor{blue}{0.8139}  & \textcolor{red}0.8036 \\
C3 & 0.6087 & 0.7239 & 0.7717 & \textcolor{blue}{0.9293} & 0.8860 & \textcolor{red}{0.9130}  & {0.7989} \\
C4 & 0.8949 & {0.9396} & 0.8233 & 0.9262 & 0.8971 & \textcolor{red}{0.9553} & \textcolor{blue}{0.9597} \\
C5 & 0.9704 & 0.9727 & 0.9707 & \textcolor{red}{0.9863} & 0.8998 & \textcolor{blue}{0.9932} & {0.9039} \\
C6 & 0.5969 & 0.7723 & 0.6906 & 0.6471 & 0.7222 & \textcolor{red}{0.8181} & \textcolor{blue}{0.9999} \\
C7 & 0.6481 & 0.5885 & 0.5356 & 0.6878 & 0.6600 & \textcolor{red}{0.7548} & \textcolor{blue}{1.0000} \\
C8 & 0.4868 & 0.5703 & 0.6306 & 0.6338 & 0.5709 & \textcolor{red}{0.7376} & \textcolor{blue}{0.9979} \\
C9 & 0.4433 & 0.7287 & 0.6507 & 0.6933 & 0.5709 & \textcolor{red}{0.7376} & \textcolor{blue}{1.0000} \\
C10 & 0.9630 & 0.9344 & \textcolor{red}{0.9506} & \textcolor{blue}{0.9877} & 0.9753 & \textcolor{blue}{0.9877} & {0.8395} \\
C11 & 0.7428 & \textcolor{blue}{1.0000} & 0.8867 & 0.8778 & 0.8762 & \textcolor{red}{0.9317} & {0.5177} \\
C12 & 0.1545 & \textcolor{blue}{0.8818} & 0.5000 & 0.5000 & 0.6394 & \textcolor{red}{0.7848} & \textcolor{red}{0.8545} \\
C13 & 0.9111 & \textcolor{blue}{1.0000} & 0.9778 & \textcolor{blue}{1.0000} & 0.9556 & \textcolor{blue}{1.0000} & \textcolor{blue}{1.0000} \\
C14 & 0.3333 & \textcolor{red}{0.8462} & 0.6667 & 0.4872 & 0.7949 & {0.7949} & \textcolor{blue}{0.7949} \\
C15 &\textcolor{blue}{1.0000} & \textcolor{blue}{1.0000} & 0.8182 & 0.7273 & {0.9091} & \textcolor{blue}{1.0000} & \textcolor{red}{0.9091} \\
C16 &0.8000 &\textcolor{blue}{1.0000} & \textcolor{blue}{1.0000}& 0.8000 & 0.8000 & \textcolor{blue}{1.0000} & \textcolor{blue}{1.0000} \\
\hline
OA &0.7043 & 0.7589 & 0.7066 & 0.7511 & 0.7186 &\textcolor{red} {0.8176}& \textcolor{blue}{0.8992}     \\
AA &0.7960 & 0.8664 &0.7637 & 0.7803 & 0.7897&\textcolor{red} {0.8781}& \textcolor{blue}{0.8982}     \\
$\kappa$ & 0.6642 & 0.7281 & 0.6673 & 0.7164 & 0.6804 &\textcolor{red} {0.7919}  & \textcolor{blue}{0.8857}  \\
\hline
\bottomrule
\end{tabular}%

\label{ip_mamba}
\end{table*}

\subsubsection{University of Pavia Data Set}
The data set from the University of Pavia serves as a testing ground to exhibit the comparative edge of our HSIMamba approach against a variety of advanced classification methods. The competing methodologies, which range from 1-D CNN to SpectralFormer, stand in comparison within Table~\ref{up_mamba}. Our HSIMamba, inclusive of a spatial processing block, consistently outperformed the other techniques, claiming the highest scores in overall accuracy (OA), average accuracy (AA), and Kappa coefficient—evidenced by our results highlighted in blue. The model not only achieved an OA of 0.9814 and an AA of 0.9769, but also excelled with a Kappa value of 0.9749. This trifecta of top-tier metrics underscores the robustness of HSIMamba in the realm of hyperspectral image classification, even more so as it took the lead across the majority of the individual classes with substantial margins.

The comprehensive quantitative performance analysis is detailed in Table~\ref{up_mamba}, which includes class-specific accuracies, further underscoring the robust and consistent performance of our methodology. While SpectralFormer showed commendable results as the runner-up in Kappa values, indicating strong performance, HSIMamba demonstrated dominance that was particularly pronounced across the classes, asserting its preeminence in hyperspectral data interpretation and classification.

\begin{table*}[!pht]
\centering
\caption{Quantitative Performance of Different Classification Methods in Terms of OA, AA, and $\kappa$, as well as the Accuracies for Each Class on the University of Pavia Data set. \textcolor{blue}{Blue Represents the Best method}, \textcolor{red}{Red Represents the second place.}}
\begin{tabular}{c|>{\centering\arraybackslash}m{1.5cm}|>{\centering\arraybackslash}m{1.5cm}|>{\centering\arraybackslash}m{1.5cm}|>{\centering\arraybackslash}m{1.5cm}|c|c|c}
\hline
\toprule
{Class No} &{1-D}  & {2-D} & RNN & {mini}&Transformer &SpectralFormer &OurMethod\\ 
& CNN & CNN &  &GCN & (ViT)&(patch-eise) & HSIMamba+SpatialBlock\\
\hline
\hline
C1  & \textcolor{red}{0.8890}&0.8098  & 0.8401 & 0.9635 &{0.7151} &0.8273 &\textcolor{blue}{0.9675}  \\
C2 & {0.5881}&0.8170 & {0.6695} & 0.8943& 0.7682 & \textcolor{red}{0.9403}& \textcolor{blue}{0.9892}  \\
C3 & {0.7311}&0.6799 & 0.5746 & \textcolor{red}{0.8701} & 0.4639& 0.7366& \textcolor{blue}{0.9532 }\\
C4  & {0.8207}& 0.9736  & \textcolor{red}{0.9770}  & 0.9426   &\textcolor{red}{0.9639}&0.9375&\textcolor{blue} {0.9836} \\
C5 &0.9946 &\textcolor{red}{0.9964} &\textcolor{red}{0.9910} & 0.9982 & 0.9919 &{0.9928}&  \textcolor{blue}{1.0000}   \\
C6 & 0.9792 &\textcolor{blue}{0.9759}  & {0.8318}  & 0.4312 & {0.8318}  &\textcolor{red}{0.9075}& \textcolor{blue}{0.9922 }  \\
C7 & {0.8807} & 0.8247 & 0.8308 & \textcolor{red}{0.9096}& {0.8308} &0.8756& \textcolor{blue}{0.9749}  \\
C8 &0.8814 &\textcolor{blue}{0.9762} & 0.8963 & 0.7742  & 0.8963  &\textcolor{red}{0.9581}& {0.9473}   \\
C9 & \textcolor{red}{0.9987}&{0.9560} & 0.9648& 0.8727& 0.9648 &0.9421& \textcolor{blue}{1.0000}   \\
\hline
OA &0.7550 & 0.8605& 0.7713& {0.7979}  & 0.7699&\textcolor{red}{0.9107}& \textcolor{blue}{0.9808}   \\
AA & 0.8626&0.8899  & 0.8429 & 0.8507   & 0.8022  &\textcolor{red}{0.9020}& \textcolor{blue}{0.9787}     \\
$\kappa$ & 0.6948&0.8187 & 0.7101& {0.7367} & 0.7010&\textcolor{red}{0.8805} &\textcolor{blue}{0.9741}   \\
\bottomrule
\hline
\end{tabular}%

\label{up_mamba}
\end{table*}

\subsection{Efficiency Analysis}
This section evaluates the computational efficiency of the HSIMamba model, focusing on memory consumption, training, and testing durations using a GPU. The analysis was conducted with the Houston 2013 dataset, emphasizing the model's performance in relation to varying patch sizes. Notably, for the patch size yielding the highest Overall Accuracy (OA), the average training time was recorded at 161 seconds, while the model completed the testing phase in just 1.19 seconds. Furthermore, the model demonstrated prudent memory usage, requiring only 126 MB during training. These metrics attest to the HSIMamba model's exceptional efficiency, balancing memory demands with swift training and testing times, thereby underscoring its viability for extensive hyperspectral image analysis tasks.

Table~\ref{eff} provides a detailed breakdown of the performance metrics, including OA, memory usage (MB), training time (seconds), and testing time (seconds) across different patch sizes. This comprehensive portrayal not only highlights the model's adaptability to various patch sizes but also reinforces its efficiency and effectiveness in handling hyperspectral data, making it a formidable tool for real-time applications and extensive datasets.

\begin{table*}[!htbp]
\centering
\caption{The Training Model Memory Used, Training Time, Test Time, OA Based on Different Patch Size of Houston 2013 Data set}
\label{eff}
\begin{tabular}{@{}lcccccccc@{}}
\hline
\toprule
Metrics   & 1      & 3      & 5      & 7      & 9      & 11     & 13     & 15     \\
\hline
\hline
OA        & 0.9383 & 0.9638 & 0.9789 & 0.9517 & 0.9561 & 0.9766 & 0.9626 & 0.9651 \\
Memory (MB) & 208.75 & 136.53 & 160.98 & 196.88 & 243.74 & 309.1  & 378.07 & 544.15 \\
Training (s) & 152.3  & 159.53 & 170    & 183.57 & 200.49 & 206.17 & 246.1  & 257.95 \\
Test (s)  & 1.07   & 1.09   & 1.19   & 1.27   & 1.39   & 1.54   & 1.65   & 1.88   \\
\bottomrule
\hline
\end{tabular}
\end{table*}

\section{Conclusions}
This research has introduced HSIMamba, an innovative bidirectional hyperspectral imaging model that seamlessly incorporates bidirectional state space modules into its design. This model is distinguished by its ability to master the dual challenges of compressing high-dimensional spectral data and navigating the intricacies of both spatial and spectral dimensions with unprecedented efficiency.

Extensive evaluations on three different hyperspectral datasets have consistently shown that HSIMamba outperforms traditional transformer-based approaches. This significant advance not only underscores the robustness and flexibility of the model, but also heralds a paradigm shift in hyperspectral image classification.

In contrast to the resource-intensive nature of conventional transformer models, a key feature of HSIMamba is its exceptional efficiency. HSIMamba has been specifically designed to reduce memory requirements, enabling rapid training and testing without sacrificing accuracy. This efficiency allows advanced hyperspectral image analysis to be deployed on platforms with limited computing capacity, thereby democratising access to cutting-edge remote sensing technologies.

In addition, the frugal use of resources by the model creates opportunities for real-time applications and the management of large datasets in resource-limited environments. The combination of efficiency and improved performance makes HSIMamba a crucial advancement in hyperspectral imaging, offering a future where advanced remote sensing technologies are both powerful and accessible.

In summary, the introduction of HSIMamba is a significant advancement in hyperspectral imaging. Its unique combination of bidirectional processing, outstanding classification performance, and computational efficiency sets new standards for the field. This model not only showcases the potential of incorporating state-space models into hyperspectral analysis but also encourages further exploration into their application in remote sensing data analysis. The results of this study support the innovative trajectory of HSIMamba and suggest further research into its capabilities and broader implications for remote sensing. 

\bibliographystyle{plain} 
\bibliography{HSIMamba}

\vfill

\end{document}